\documentclass[conference]{IEEEtran}
\IEEEoverridecommandlockouts
\usepackage{cite}
\usepackage{amsmath,amssymb,amsfonts}
\usepackage{algorithmic}
\usepackage{graphicx}
\usepackage{textcomp}

\usepackage{multirow}
\usepackage{tabularx,booktabs}
\newcolumntype{C}{>{\centering\arraybackslash}X} 
\usepackage[dvipsnames]{xcolor}
\usepackage{longtable}
\usepackage{makecell, cellspace, caption}
\usepackage{array}
\usepackage{subcaption}
\usepackage{cuted}
\usepackage{multirow}
\usepackage[utf8]{inputenc}
\usepackage{tabularx}
\usepackage{array}
\usepackage{amssymb}
\usepackage{makecell}
\usepackage[margin=2cm]{geometry}
\usepackage{adjustbox}
\usepackage{multirow}
\usepackage{tabulary,booktabs}
\usepackage{ragged2e}
\usepackage{amsmath}
\usepackage{booktabs}
\usepackage{array}
\newcolumntype{L}{>{$}l<{$}}
\newcolumntype{C}{>{$}c<{$}}
\newcolumntype{R}{>{$}r<{$}}

\usepackage{amssymb}
\usepackage{pifont}
\usepackage{graphicx}
\usepackage[caption=false]{subfig}
\usepackage{url}
\usepackage{flushend}

\def\BibTeX{{\rm B\kern-.05em{\sc i\kern-.025em b}\kern-.08em
    T\kern-.1667em\lower.7ex\hbox{E}\kern-.125emX}}
\begin{document}

\title{SINET: Sparsity-driven Interpretable Neural Network for Underwater Image Enhancement}

\author{\IEEEauthorblockN{Gargi Panda}
\IEEEauthorblockA{\textit{Department of Electrical Engineering} \\
\textit{IIT Kharagpur, India}\\
pandagargi@gmail.com}
\and
\IEEEauthorblockN{Soumitra Kundu}
\IEEEauthorblockA{\textit{Rekhi Centre of Excellence for the Science of Happiness} \\
\textit{IIT Kharagpur, India}\\
soumitra2012.kbc@gmail.com}
\and
\IEEEauthorblockN{Saumik Bhattacharya}
\IEEEauthorblockA{\textit{Department of Electronics and Electrical Communication Engineering} \\
\textit{IIT Kharagpur, India}\\
saumik@ece.iitkgp.ac.in}
\and\IEEEauthorblockN{Aurobinda Routray}
\IEEEauthorblockA{\textit{Department of Electrical Engineering} \\
\textit{IIT Kharagpur, India}\\
aroutray@ee.iitkgp.ac.in}
}

\maketitle

\begin{abstract}
Improving the quality of underwater images is essential for advancing marine research and technology. This work introduces a sparsity-driven interpretable neural network (SINET) for the underwater image enhancement (UIE) task. Unlike pure deep learning methods, our network architecture is based on a novel channel-specific convolutional sparse coding (CCSC) model, ensuring good interpretability of the underlying image enhancement process.
The key feature of SINET is that it estimates the salient features from the three color channels using three sparse feature estimation blocks (SFEBs). The architecture of SFEB is designed by unrolling an iterative algorithm for solving the $\ell_1$ regulaized convolutional sparse coding (CSC) problem. Our experiments show that SINET surpasses state-of-the-art PSNR value by $1.05$ dB with  $3873$ times lower computational complexity. 
Code can be found at: \url{https://github.com/gargi884/SINET-UIE/tree/main}.

\end{abstract}
\begin{figure*}[hbt!] 
    \centering
  {\includegraphics[width=0.8\linewidth]{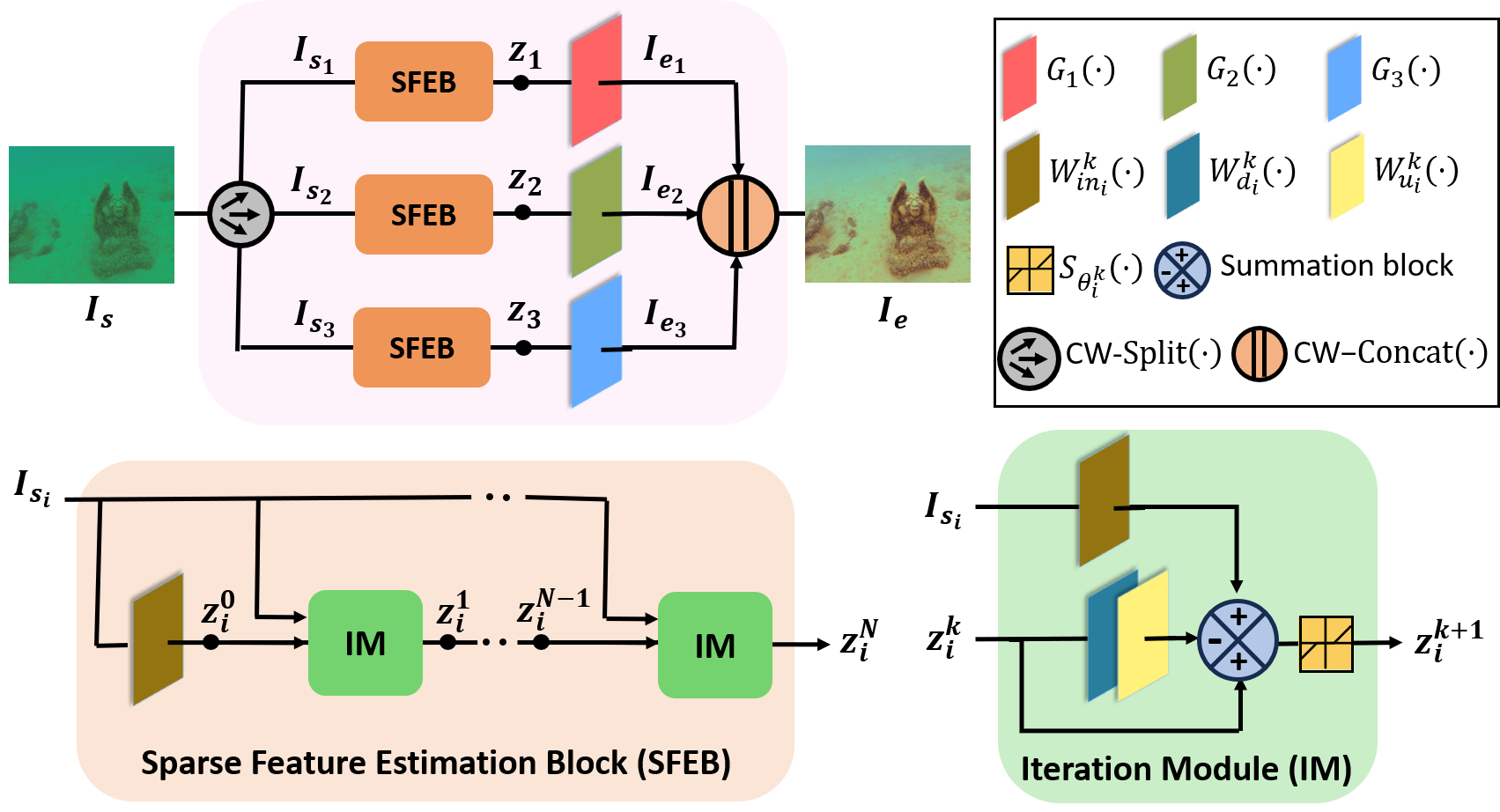}}
  \caption{Network architecture of SINET and the structure of SFEB.}
  \label{fig1} 
\end{figure*}
\section{Introduction}
\label{sec:intro}
Enhancing the quality of underwater images is essential for performing various computer vision applications in marine biology, underwater archaeology, and robotics. However, the inherent challenges of underwater environments, including uneven scattering, refraction, and non-uniform color channel attenuation, significantly degrade image quality and distort the true appearance of underwater scenes. To address these challenges and enhance the usability of underwater images for high-level vision tasks, numerous underwater image enhancement (UIE) techniques \cite{uieb,uiewd,scnet,light,deepwavenet,x_caunet,DSFFNet} have been developed. Some methods, such as \cite{deepwavenet,x_caunet}, employ deep neural networks (DNNs) specifically designed for UIE, considering the non-uniform attenuation of different color wavelengths. Compared to other methods, these color channel-specific designs have proven to be highly effective for UIE. However, despite the successes of deep learning (DL) approaches in learning complex mappings between degraded and enhanced images, they often fall short in terms of interpretability. The black-box nature of DL models obscures the understanding of how different color channels behave during the enhancement process. This limitation highlights the need for UIE techniques that not only enhance image quality but also provide clear insights into the enhancement mechanisms. 

The convolutional sparse coding (CSC) model is widely adopted in image processing due to its interpretability and robust theoretical framework \cite{csc_cvpr,sparse}. With sparsity constraints, CSC can effectively extract meaningful structures from images. Given an image $I\in\mathbb{R}^{H \times W \times 3}$, CSC expresses $I$ using convolutional sparse representations $z\in\mathbb{R}^{H \times W \times K}$,

\begin{equation}\label{eq1}
    I = D(z)
\end{equation}

\noindent
where $D(\cdot)$ is a convolutional dictionary which can be learned and $K$ is the number of convolution filters. $H$ and $W$ are the image height and width, respectively. In Eqn. (\ref{eq1}), $z$ is enforced to be sparse by $\ell_0$ or $\ell_1$ norm regularization, which helps to highlight the salient features in the input image $I$. Optimization methods like the alternating direction method of multipliers (ADMM) \cite{admm} are prevailing approaches for solving Eqn. (\ref{eq1}). However, such algorithms are time-consuming, limiting their large-scale data applications.

To incorporate the advantages of optimization and deep learning methods, algorithm unrolling-based methods have been developed \cite{unrolling, urolling_boyd}. These methods unroll the iteration steps of an optimization problem into a deep neural network. CSC-based algorithm unrolling networks have been applied in several image processing tasks like image denoising \cite{ref1, denoising} and super-resolution \cite{ref2, sr}. Since the CSC-based unrolling network provides an interpretable and efficient framework, it can also be effective for UIE. However, for UIE, addressing the unique challenge of wavelength-dependent light attenuation is important. In this context, employing a separate CSC model for each color channel can be highly effective.

To address the UIE problem, this work proposes a novel color channel-specific convolutional sparse coding (CCSC) model, based on which we design a sparsity-driven interpretable neural network (SINET). Such a model-based design provides good interpretability of the underlying image enhancement process in SINET.
Specifically, we propose an algorithm unrolling-based sparse feature estimation block (SFEB) to estimate the salient features from the different color channels. Experiments across two benchmark datasets show that, with very low computational complexity, SINET surpasses the state-of-the-art methods.

The remaining paper is organized as follows. Section \ref{sec:method} describes our proposed CCSC model, the architectures of SFEB and SINET, and the loss function in detail. In Section \ref{sec:experiments}, we present extensive experiments to evaluate our method. Finally, Section \ref{sec:conclusion} concludes the work. 
\section{Proposed Method}
\label{sec:method}
\subsection{Problem Formulation}

Underwater images experience degradation due to the uneven scattering and absorption of light. Moreover, this degradation varies across the color channels, as each wavelength undergoes a different level of attenuation \cite{deepwavenet,x_caunet}. Based on this observation, we represent the three color channels with three separate models. We consider the degraded image $I_s \in \mathbb{R}^{H \times W \times 3}$, which  we split into three color channels: $I_{s_i} \in \mathbb{R}^{H \times W \times 1}, i\in\{1,2,3\}$. $H$ and $W$ are image height and width, respectively. Each channel is expressed as,
\begin{equation} \label{eq_ccsc}
\begin{split}
&I_{s_1},I_{s_2},I_{s_3} = \text{CW-Split}(I_s)\\
&I_{s_i} = D_i(z_i)
\end{split}
\end{equation}

\noindent
where $\text{CW-Split}(\cdot)$ is channel-wise splitting operation, $D_i(\cdot)$ is convolution operation which can be learned, and $z_i\in \mathbb{R}^{H \times W \times K}$ is the salient feature in the input channel. $K$ is the number of convolution filters. 

Then, from the salient feature $z_i$, we restore the enhanced color channel
$I_{e_i} \in \mathbb{R}^{H \times W \times 1}$ using convolution operation. Finally,
we channel-wise concatenate $I_{e_i}$, $i\in\{1,2,3\}$ to obtain the enhanced image $I_e \in \mathbb{R}^{H \times W \times 3}$. The restoration process can be expressed as,

\begin{equation} \label{eq_restoration}
\begin{split}
&I_{e_i} = G_i(z_i)\\
&I_e = \text{CW-Concat}(I_{e_1},I_{e_2},I_{e_3})
\end{split}
\end{equation}

\noindent
where $\text{CW-Concat}(\cdot)$ is channel-wise concatenation operation and $G_i(\cdot)$ is convolution operation. 

In Eqn. (\ref{eq_ccsc}), we constrain $z_i$ to be $\ell_1$ regularized. The $\ell_1$ regularization is a widely used technique that promotes sparsity in the features and thus highlights the salient features. So, we estimate $z_i$ by solving the following Eqn.,
\begin{equation}\label{l1}
\underset{z_i}{\mathrm{Argmin}}\:\frac{1}{2}\Big|\Big|\:I_{s_i}-D(z_i)\Big|\Big|_2^2 +\lambda ||z_i||_1 
\end{equation}

\noindent
Eqn. (\ref{l1}) is an $\ell_1$ regularized CSC problem, for solving which we propose a sparse feature estimation block described in the next subsection.
\begin{figure*}[hbt!] 
    \centering
  {\includegraphics[width=\linewidth]{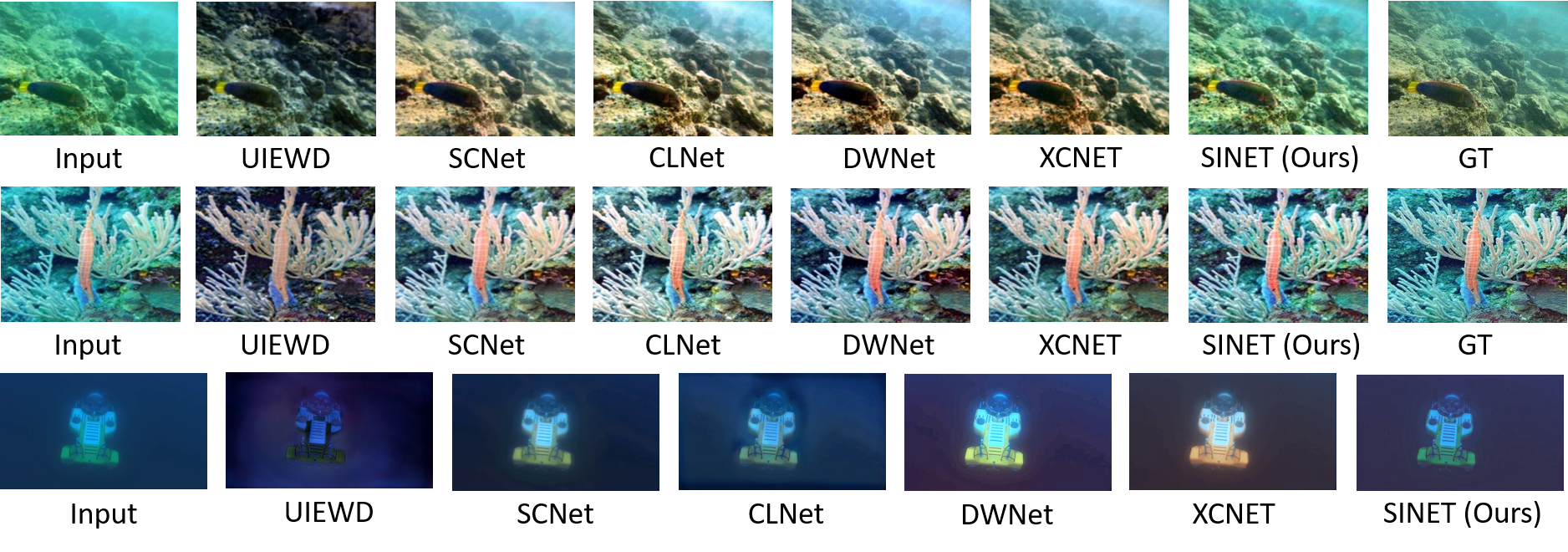}}
  \caption{Visual comparison with SOTA methods. First two rows: Images from the LSUI dataset. Third row: Image from the UIEBC dataset, which has no ground truth (GT) image. Images are best viewed in $400\%$ zoom.}
  \label{fig2} 
\end{figure*}
\subsection{Sparse Feature Estimation Block (SFEB)}

In SFEB, we estimate the sparse features by solving Eqn. (\ref{l1}). Sreter \textit{et al.} \cite{lcsc} proposed a convolutional extension of the learned iterative shrinkage thresholding algorithm (LISTA) \cite{lista} to solve the $\ell_1$ regularized CSC problem. From \cite{lcsc}, the iteration step for updating $z_i$ is,

\begin{equation}\label{eq_ista1}
z_i^{k+1}=S_{\theta_i}\Big(z_i^k- W_{u_i}\Big(W_{d_i}\Big(z_i^k\Big)\Big)+W_{u_i}\Big(I_{s_i}\Big)\Big)
\end{equation}

\noindent
where $z_i^k$ is the estimation of $z_i$ at $k^{th}$ iteration, $W_{u_i}(\cdot)$ and $W_{d_i}(\cdot)$ are learnable convolution layers. $S_{\theta_i}(\cdot)$ is the soft thresholding function defined as,

\begin{equation}\label{soft}
S_{\theta_i}(x)=sgn(x)\:max(|x|-{\theta_i}, 0)    
\end{equation}

\noindent
$sgn(\cdot)$ is the sign function. $\theta_i$ is the threshold value, which is a learnable parameter. As shown in Eqn. (\ref{soft}), $S_{\theta_i}(\cdot)$ nullifies the input values with an absolute values lower than $\theta_i$. Thus, it promotes sparsity and highlights the salient features in the input color channel. 

However, in Eqn. (\ref{eq_ista1}), the learnable parameter $\theta_i$ and convolution layers $W_{u_i}(\cdot)$, $W_{d_i}(\cdot)$ are shared across all the iterations in \cite{lcsc}. Such parameter sharing limits the performance of the sparse estimation. Motivated by \cite{maximal}, we use different $\theta_i$ and different convolution layers for each iteration to improve the estimation accuracy. Additionally, for the input $I_{s_i}$, using a learnable layer that is different from $W_{u_i}(\cdot)$ can have better performance \cite{adalista}. Incorporating these modifications in Eqn. (\ref{eq_ista1}), the iteration step for updating $z_i$ becomes,

\begin{equation}\label{eq_ista2}
z_i^{k+1}=S_{\theta_i^k}\Big(z_i^k- W_{u_i}^k\Big(W_{d_i}^k\Big(z_i^k\Big)\Big)+W_{in_i}^k\Big(I_{s_i}\Big)\Big)
\end{equation}

\noindent
Moreover, in \cite{lcsc}, the learnable parameter $\theta_i$ is not constrained to be positive and can learn any value. A negative value of $\theta_i$ contradicts its definition. Also, $\theta_i$ should decrease with iteration number $k$ because the image degradation level reduces with iterations. Inspired by \cite{fistanet}, we constrain $\theta_i^k$ as,

\begin{equation} \label{eq_constrain}
\begin{split}
&\theta_i^{k}=sp(w_{\theta_i} k+b_{\theta_i})\:\:,\:\: w_{\theta_i} <0
\end{split}
\end{equation}

\noindent
where $sp(\cdot)$ is the softplus function, $w_{\theta_i}$ and $b_{\theta_i}$ are learnable parameters. We unfold Eqn. (\ref{eq_ista2}) into the iteration module (IM) shown in Fig. \ref{fig1}. Assuming $z_i^k=0; \:\: \forall k<0$ and stacking a number of iteration modules, we construct our sparse feature estimation block (SFEB).
\subsection{Architecture of SINET}
\label{MINet}

Fig. \ref{fig1} shows the architecture of SINET. Given the source image $I_s$, first the source color channels $I_{s_1}$, $I_{s_2}$ and $I_{s_3}$ are obtained using the channel splitting operation. Then, using three SFEBs, SINET estimates the salient features $z_1$, $z_2$, and $z_3$. Now, using the convolution layers $G_1(\cdot)$, $G_2(\cdot)$, and $G_3(\cdot)$, we get the three enhanced color channels $I_{e_1}$, $I_{e_2}$ and $I_{e_3}$ respectively. Finally, we obtain the enhanced image $I_e$ by concatenating the three reconstructed color channels.
\subsection{Loss Function}
\label{loss}
During the training of SINET, the enhanced image $I_e$ is constrained to be similar to the ground truth image $I_g$. We constrain the intensity, texture, and structural similarities between $I_e$ and $I_g$. For intensity similarity, we use the $L_1$ loss function,
\begin{equation}\label{int}
\mathcal{L}_{int}=||I_e-I_g||_1
\end{equation}

\noindent
To maintain texture similarity, we use the $L_1$ loss function between the image gradients,
\begin{equation}\label{text}
\mathcal{L}_{text}=||\nabla I_e-\nabla I_g||_1
\end{equation}

\noindent
where $\nabla$ is the Sobel gradient operator. For structural similarity, we use the following loss function,
\begin{equation}\label{ssim}
\mathcal{L}_{ssim}=1-ssim(I_e,I_g)
\end{equation}

\noindent
$ssim(\cdot)$ measures the structural similarity between two images. The overall loss function becomes,
\begin{equation}\label{loss}
\mathcal{L} = \alpha_1  \mathcal{L}_{int} +\alpha_2 \mathcal{L}_{text} +\alpha_3 \mathcal{L}_{ssim}
\end{equation}

\noindent
$\alpha_1$, $\alpha_2$ and $\alpha_3$ are the weights for different loss components. Experimentally we set $\alpha_1=40$, $\alpha_2=40$, $\alpha_3=100$.
\section{Experiments}
\label{sec:experiments}
\subsection{Setup}

We utilize publicly available datasets for training and testing SINET. For training, we use 890 images from the UIEB \cite{uieb} dataset. We train SINET using Adam optimizer with a constant learning rate of $1\times10^{-4}$ and batch size of $4$. While training, the images are randomly cropped to the resolution of $256\times 256$. For image augmentation, we utilize horizontal and vertical flipping. We test the performance on 60 images from the LSUI \cite{lsui} dataset without finetuning to check the model's generalization ability. Moreover, to check the model's performance on real-world images, we test on 60 images from the UIEB-challenge (UIEBC) dataset, which has no ground truth images. We compare the PSNR and SSIM metrics for the LSUI dataset, which has ground truth images. For the UIEBC dataset, we compare the performance using the UIQM metric \cite{uiqm}. For the implementation of SINET, we set the kernel size for each convolutional layer as $11\times11$ and the number of filters as $K=16$. The number of iteration modules in SFEB is set to $4$. All experiments are conducted within the Pytorch framework using an NVIDIA A40 GPU. 
A short demo of our work is available at: \url{https://github.com/gargi884/SINET-UIE/blob/main/figs/demo.gif}.
\subsection{Performance Comparison}
We compare SINET with five state-of-the-art methods: UIEWD \cite{uiewd}, SCNet \cite{scnet}, CLNet \cite{CLUIENet}, DWNet \cite{deepwavenet}, and XCNET \cite{x_caunet}. Figure \ref{fig2} shows the visual comparison results on two images from the LSUI dataset and one image from the UIEBC dataset. UIEWD produces severe color distortions. For the images from the LSUI dataset, SCNet, CLNet, DWNet, and XCNET change the greenish tone that is present in the GT image. Whereas SINET preserves the greenish tone and reconstructs the images with better structure similarity with the GT image. For the image from the UIEBC dataset, SCNet, DWNet, and XCNET change the green portion of the input image to a different color. Though CLNet preserves the green color, but it produces a blurry image. In contrast, SINET preserves the green color and produces an image with better structural details. 
More visual comparisons can be found at: \url{https://github.com/gargi884/SINET-UIE/blob/main/figs/visual.pdf}.

Table \ref{tab:main1} presents the quantitative comparison. Along with the PSNR, SSIM, and UIQM metrics, we also list the FLOPs, which are calculated under the setting of enhancing an underwater image of resolution $640\times320$. As the results show, SINET surpasses other methods with the lowest computational complexity. Specifically, XCNET has comparable performance with SINET, but our method has $3873$ times lower computational complexity. SINET achieves a $1.05$ dB gain of PSNR on the LSUI dataset. All these results show the effectiveness of our method. 
\begin{table}[hbt]
\fontsize{9.1}{11.1}\selectfont
\centering
\caption{Comparison with SOTA methods on two benchmark datasets. FLOPs are calculated under the setting of enhancing an underwater image of resolution $640\times 320$. A low value of FLOPs and a high value of PSNR, SSIM, and UIQM are desired. The best values are highlighted in \textcolor{red}{\textbf{red}} and the second-best values are underlined in \textcolor{blue}{\underline{blue}} color.
}
\begin{tabular}{lcccc}
\toprule
\multirow{2}{*}{Method} & \multirow{2}{*}{FLOPs(G)} &
\multicolumn{2}{c}{LSUI \cite{lsui}} &
\multicolumn{1}{l}{UIEBC \cite{uieb}}\\
\cmidrule(lr){3-4}
\cmidrule(lr){5-5}
& &
\multicolumn{1}{c}{PSNR}&
\multicolumn{1}{c}{SSIM}& \multicolumn{1}{c}{UIQM}\\
\midrule
\text{UIEWD (2022)} & 160.12 & 16.37   & 0.6634 & 2.421 \\
\text{SCNet (2022)} & 75.86 & 23.01   & 0.8062 & 2.681\\
\text{CLNet (2023)} & 96.84 & 20.37   & 0.7797 & \textcolor{blue}{\underline{2.693}} \\
\text{DWNet (2023)} & \textcolor{blue}{\underline{56.62}} & 22.04   & 0.7828 & 2.434 \\
\multirow{1}{*}{XCNET (2024)} & 193.66 & \textcolor{blue}{\underline{23.08}} & \textcolor{blue}{\underline{0.8080}} & \textcolor{red}{\textbf{2.713}} \\
\multirow{1}{*}{SINET (Ours)}  & \textcolor{red}{\textbf{0.05}} & \textcolor{red}{\textbf{24.13}} & \textcolor{red}{\textbf{0.8086}} & \textcolor{red}{\textbf{2.713}}\\
\bottomrule
\end{tabular}
\label{tab:main1}
\end{table}
\subsection{Visualization of Intermediate Features}
\begin{figure}[hbt!] 
    \centering
  {\includegraphics[width=\linewidth]{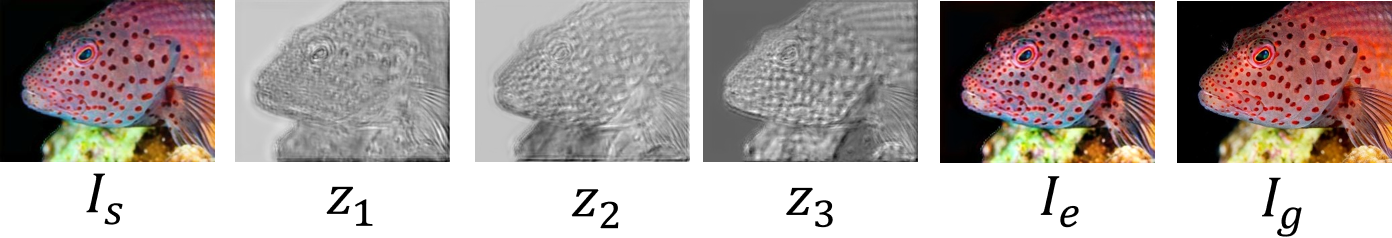}}
  \caption{Visualization of intermediate features. }
  \label{fig3} 
\end{figure}

In our SINET, the SFEBs estimate the sparse features from the three color channels. To verify this ability, Fig. \ref{fig3} shows the visualizations of sparse features for each color channel for an image from the LSUI dataset. The sparse features are different for the three color channels, and they capture the salient features in the input image.
\subsection{Ablation Experiments}
\begin{figure}[htb]
  \begin{subfigure}[t]{0.495\columnwidth}
    \includegraphics[width=\linewidth]{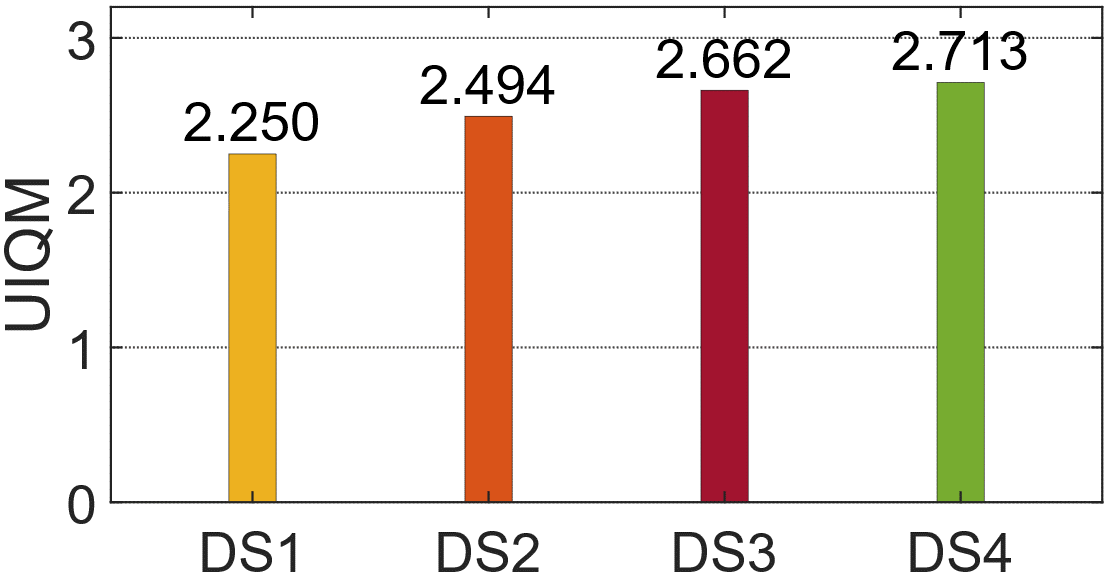}
    \caption{Design Settings (DS)}
    \label{fig:1}
  \end{subfigure}
  \hfill 
  \begin{subfigure}[t]{0.495\columnwidth}
    \includegraphics[width=\linewidth]{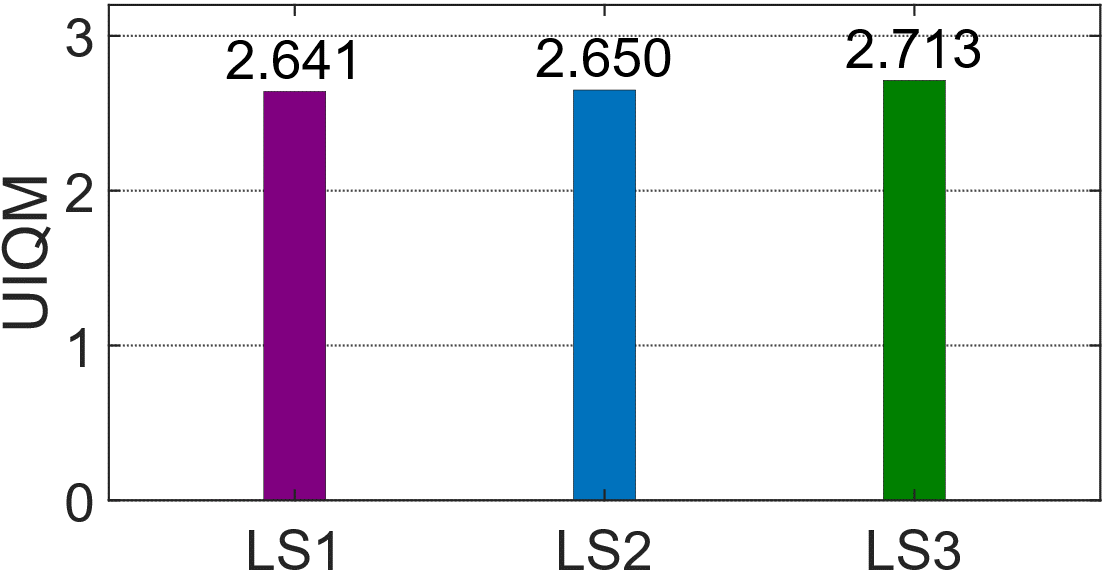}
    \caption{Loss Settings (LS)}
    \label{fig:2}
  \end{subfigure}
  \caption{Ablation experiments with different settings.}
  \label{fig4}
\end{figure}

We conduct several ablation experiments to validate the effectiveness of our method. We train the models with the UIEB dataset and test on the UIEBC dataset. To verify the design of SINET, we experiment with the following design settings (DS), \textbf{DS1:} Instead of using SFEB, use four convolution layers with kernel size $11\times11$ and filter size $16$. \textbf{DS2:} Instead of using SFEB, use the learned convolutional sparse coding block proposed in \cite{lcsc}.  \textbf{DS3:} Instead of using separate SFEBs for the three color channels, use a single SFEB and reconstruct the image as a whole. \textbf{DS4:} The proposed SINET. To evaluate the effectiveness of our loss function, we train SINET with the following loss settings (LS),  \textbf{LS1:} $\mathcal{L}_{int}$ loss only, \textbf{LS2:} $\mathcal{L}_{int}$ and $\mathcal{L}_{text}$ loss, \textbf{LS3:} $\mathcal{L}_{int}$, $\mathcal{L}_{text}$, and $\mathcal{L}_{ssim}$ loss. Fig. \ref{fig4} shows the performance for different ablation experiments and justifies our design of SINET and the loss function.

\section{Conclusion}
\label{sec:conclusion}

This work introduces SINET, a sparsity-driven interpretable neural network for the UIE task. Different from other methods, our network is designed based on a novel channel-specific CSC model. This model-based design provides good interpretability of the underlying image enhancement process. Our ablation studies justify the design of SINET. Experimental results across two benchmark datasets demonstrate that with very low computational complexity, SINET surpasses state-of-the-art methods. Future research could explore model-based interpretable networks that further incorporate the unique properties of water and light interactions in the underwater environment.

\clearpage
\bibliographystyle{IEEEtran}
\bibliography{ref}

\end{document}